%% file: main.tex
\documentclass[letterpaper, 10 pt, conference]{ieeeconf}
\IEEEoverridecommandlockouts
\usepackage{amsmath,amssymb,amsfonts}
\usepackage{algorithmic}
\usepackage{algorithm}
\usepackage{graphicx}
\usepackage{textcomp}
\usepackage[table,xcdraw]{xcolor}
\usepackage[percent]{overpic}
\usepackage{comment}
\usepackage[noadjust]{cite}

\usepackage{caption}
\usepackage{subcaption}
\usepackage{float}
\usepackage{array}
\usepackage{soul}
\usepackage{svg}
\newcommand{\ssymbol}[1]{^{\@fnsymbol{#1}}}

\usepackage[hidelinks]{hyperref}
\usepackage[nocomma]{optidef}

\title{\LARGE \bf Digital Model-Driven Genetic Algorithm for Optimizing Layout and Task Allocation in Human-Robot Collaborative Assemblies}

\author{Christian Cella$^{1}$, Matteo Bruce Robin$^{1}$, Marco Faroni$^{1}$, Andrea Maria Zanchettin$^{1}$, Paolo Rocco$^{1}$   
\thanks{
        This study was carried out within the MICS (Made in Italy – Circular and Sustainable) Extended Partnership and received funding from Next-Generation EU (Italian PNRR – M4 C2, Invest 1.3 – D.D. 1551.11-10-2022, PE00000004). CUP MICS D43C22003120001
}
\thanks{$^{1}$The Authors are with Politecnico di Milano, Dipartimento di Elettronica, Informazione e Bioingegneria, Piazza Leonardo da Vinci 32, 20133, Milano (Italy). 
        {e-mail: \tt\scriptsize \{christian.cella, matteobruce.robin, marco.faroni, andreamaria.zanchettin, paolo.rocco\}@polimi.it}. }      
}

\DeclareMathOperator*{\argmin}{argmin}

\newtheorem{assumption}{\textbf{Assumption}}

\begin{document}

\maketitle
\thispagestyle{empty}
\pagestyle{empty}

\setlength{\abovedisplayskip}{3pt}
\setlength{\belowdisplayskip}{3pt}

\begin{abstract}
This paper addresses the optimization of human-robot collaborative work-cells before their physical deployment. Most of the times, such environments are designed based on the experience of the system integrators, often leading to sub-optimal solutions. Accurate simulators of the robotic cell, accounting for the presence of the human as well, are available today and can be used in the pre-deployment. We propose an iterative optimization scheme where a digital model of the work-cell is updated based on a genetic algorithm. The methodology focuses on the layout optimization and task allocation, encoding both the problems simultaneously in the design variables handled by the genetic algorithm, while the task scheduling problem depends on the result of the upper-level one. The final solution balances conflicting objectives in the fitness function and is validated to show the impact of the objectives with respect to a baseline, which represents possible initial choices selected based on the human judgment.
\end{abstract}

\input{Chapters/1.Intro}

\input{Chapters/2.StateOfTheArt}

\input{Chapters/3.ProblemStatement}

\input{Chapters/4.Methodology}

\input{Chapters/5.Experiment}

\input{Chapters/6.Conclusion}
\bibliographystyle{IEEEtran}
\bibliography{references}

\end{document}

%% file: Chapters/1.Intro.tex

\section{Introduction}
\label{sec:intro}
The industrial relevance of Human-Robot Collaborative (HRC) applications has experienced a relevant growth in the last decades, as it allows to combine the repeatability of robotic manipulators with the dexterity of the human operators. These multi-agent systems leverage the synergy resulting from the collaboration of the actors involved in the process, creating a shared  environment that improves productivity, safety, and flexibility~\cite{riedelbauch2023benchmarking}. Moreover, the advent of industrial paradigms, such as Industry 4.0 (and 5.0 more recently), has favoured the integration of new domains in the manufacturing context, leading to the generation of Cyber-Physical Systems (CPS), that revolve around the virtualization of a physical unit into its digital counterpart, usually referred to as \textit{digital twin}~\cite{khalid2017towards}.\\ 
When it comes to increasing the throughput and reducing overall cycle times for collaborative working units, the design and sub-division of tasks is clearly crucial~\cite{dawande2007throughput}.
The generation of HRC applications is a process that requires a common pipeline, starting from the rough \textit{design} and culminating with the real deployment of the physical unit in the \textit{execution} stage.
However, it turns out that existing research typically focuses on enhancing the efficiency of physical collaborative cells, generated only based on the experience of the system integrators, rather than optimizing them from the design phase~\cite{malik2021digital}. The main reason for this is probably the reciprocal dependency of the variables involved in the sub-problems characterizing HRC applications, namely \textit{layout optimization} (LO), \textit{task allocation} (TA),  \textit{task scheduling} (TS), \textit{motion planning} (MP) and \textit{safety implementation} (SI).\\
We believe instead that the optimization of HRC applications should be tackled already from the early stages, to effectively maximize the productivity of the cell, while guaranteeing safety. In this work we propose a holistic framework conceived for optimizing a complete HRC assembly operation. The genetic algorithm at the core of the procedure allows to account for a set of Key-Performance Indicators (KPIs), obtained recursively as the result of simulations that substitute the evaluation of the \textit{black-box} function describing the complete process.\\
To the best of our knowledge, implementing a non-heuristic pipeline to be applied in the \textit{pre-deployment} phase and concerning the simultaneous optimization of the interconnected sub-problems defining a HRC application has not yet been addressed in the literature.\\
The rest of this work is organised as follows. In Section~\ref{sec:state_art} a review of the related works is presented and the main contributions are emphasized. Section~\ref{sec:problem_statement} contains the mathematical formalization of the problem under exam, while in Section~\ref{sec:methodology} the algorithms we implemented are explained in detail. Section~\ref{sec:case_study} contains the validation of the procedure presented. Finally, Section~\ref{sec:conclusions} contains the conclusions and the future works.

\begin{figure}
\centering
\includegraphics[width=0.43\textwidth]{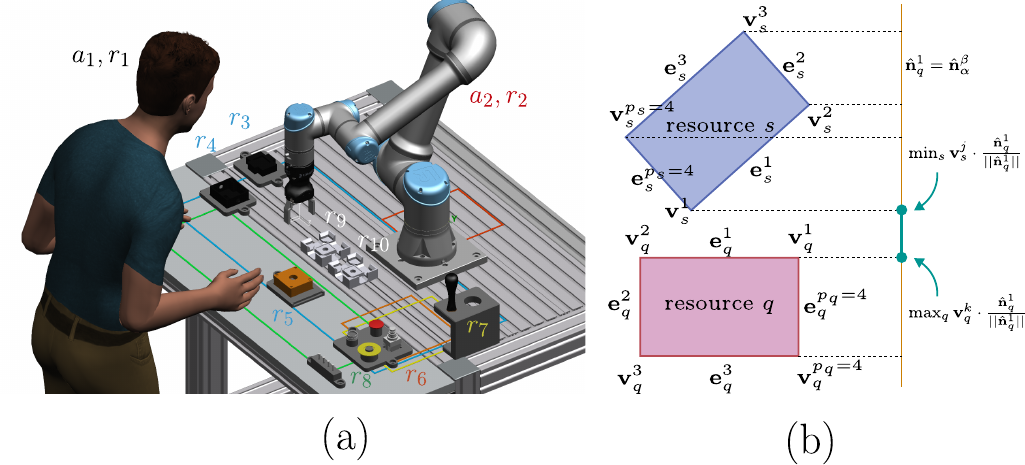}
\caption{(a) Final layout: $r_1$, $r_9$ and $r_{10}$ are fixed in the use case, while the other resources can be moved on the table. (b) Non-overlapping condition for the resources: in this case, the set $\bar{\mathcal{N}}=\{\hat{\textbf{n}}_q^1, \hat{\textbf{n}}_q^3\}$, therefore the constraint (\ref{equation:upper_level_con6}) is verified. \vspace{-0.5 cm}}
\label{fig:optimal_layout}
\end{figure}

%% file: Chapters/2.StateOfTheArt.tex
\section{Related works and contributions}
\label{sec:state_art}
The holistic optimization of HRC systems is a very challenging task, since it deals with intertwined sub-problems~\cite{5354155},~\cite{rega2021collaborative} and the mathematical relationship between them is usually unknown. Most of the works in the literature focus on improving the efficiency of existing HRC work-cells, engineered to be the solution of only one of the specific domains characterizing the collaboration~\cite{lietaert2019model}, according to one or more \textit{systematic} design methodologies~\cite{malmqvist1996comparative},~\cite{jensen2010design}.\\
A possible approach is to devise optimization algorithms accounting for a limited number of KPIs that characterize specifically one of the domains. As an example,~\cite{tay1996optimising} aims at generating the layout of a robotized work-cell by minimizing the path followed by the robot during the operations, enforcing a matrix representation of the resources. Instead of relying on a heuristic framework, many different works in the literature exploit \textit{evolutionary} algorithms to sample the search space effectively. For instance, in~\cite{lim2016nature} a combination of five \textit{nature-inspired} algorithms is proposed to find the optimal disposition of the set of resources under exam, leveraging the \textit{sequence-pair} method as the representation scheme. Similarly, the work presented in~\cite{eswaran2024optimal} relies on the capabilities of a modified \textit{Particle Swarm Optimization} (PSO) algorithm to find the optimal layout of pieces after the tasks have been allocated to the agents based on ergonomics and tolerated payload. Another approach, dealing with task allocation and layout, is shown in~\cite{ma2024development}, where a genetic algorithm is devised to find the optimal placement of the resources in a collaborative assembly line with the aim of minimizing the cycle time.\\ 
Favoured by the push of the new industrial paradigms, the problem presented above can be tackled by leveraging the capabilities of simulators, that allow to investigate different \textit{what-if} scenarios~\cite{kousi2021digital}. In~\cite{tsarouchi2016decision} the parameters of the simulation-based framework for the layout generation are constantly changed by the operator that interacts at different stages with a graphical viewer produced by the algorithm. In~\cite{raza2021pdca}, the authors introduce an heuristic human-centered methodology, that relies on the human feedback to adjust the set of parameters used in the simulations. Another example is the procedure described in~\cite{ore2019simulation}, which presents a workflow for evaluating different candidate layouts for HRC working units: if the suggestion does not meet the requirements, human designers must adjust design variables by directly intervening in the process.\\
From the analysis presented, there seems to be a significant lack of procedures that can optimize a HRC application from the initial design stages. With respect to the previous works, the main contributions of our methodology are the following:\\
1) We propose a simulation-based recursive optimization framework to be applied in the pre-deployment phase. In this stage, the physical working unit is yet to be deployed: to overcome the lack of data, we substitute the evaluation of the black-box function describing the overall process with the iterative simulations executed on a digital model, that is driven by a genetic algorithm. Unlike the method presented in~\cite{raza2021pdca} and~\cite{ore2019simulation}, our framework does not require the intervention of the human designer at any stage of the process.\\
2) In contrast with~\cite{lim2016nature} and~\cite{eswaran2024optimal}, where the optimization is not related to the different phases of the HRC process but just to one of the sub-problems, we solve the problem accounting for layout, task allocation, scheduling and motion planning, after introducing reasonable assumptions related to the process.\\
3) Our formulation of the HRC assembly process is general and not engineered for a specific use-case; the framework we propose would not lose validity if the variables or the KPIs involved in the process were changed.


%% file: Chapters/3.ProblemStatement.tex
\section{Problem statement and formalization}
\label{sec:problem_statement}
Optimizing a HRC application at the design stage entails the resolution of nested sub-problems, according to the \textit{leader-follower} paradigm presented in~\cite{du2019review},~\cite{dogan2023bilevel}. The position of the resources (LO) directly influences which of the agents can perform an operation and the time required to execute it (TA-TS); moreover, LO, TA, TS directly impact the planning phase (MP), responsible for generating collision-free trajectories and implementing safety measures (SI), according to ISO-TS/15066~\cite{ISOTS15066}. The resolution of the problem is based on the idea that the \textit{leader} makes a decision first, while the \textit{follower} reacts to that choice. In the case of HRC applications, assuming that TA and TS are encoded in the same problem (TAS) and safety is embedded in the motion planning phase (MPS), this paradigm is repeated at two different levels: first, LO is the leader, while the follower is the resolution of TAS; in the second level, TAS becomes the leader, while MPS the follower. The general problem is the following:
\begin{mini!}|s|[2]
    {\textbf{x}_{\text{LO}}} 
    {\textbf{f}_{\text{LO}}(\textbf{x}) \label{equation:general_upper}} 
    {\label{equation:general_complete_equation}} 
    {} 
    \addConstraint{\textbf{g}_{\text{LO}}(\textbf{x}) \leq \textbf{0}}{\label{equation:general_con1}} 
    \addConstraint{\textbf{x}_{\text{TAS}} \in \argmin_{\textbf{x}_{\text{TAS}}} \textbf{f}_{\text{TAS}}(\textbf{x})}{\label{equation:general_middle}} 
    \addConstraint{\text{s.t.}\ \textbf{g}_{\text{TAS}}(\textbf{x}) \leq \textbf{0}}{\label{equation:general_con2}} 
    \addConstraint{\quad\ \textbf{x}_{\text{MPS}} \in \argmin_{\textbf{x}_{\text{MPS}}} \textbf{f}_{\text{MPS}}(\textbf{x})}{\label{equation:general_lower}} 
    \addConstraint{\quad\ \text{s.t.} \quad \textbf{g}_{\text{MPS}}(\textbf{x}) \leq \textbf{0}}{\label{equation:general_con3}} 
\end{mini!}

\noindent where $\textbf{x}=[\textbf{x}_{\text{LO}}^T, \textbf{x}_{\text{TAS}}^T, \textbf{x}_{\text{MPS}}^T]^T\in\mathbb{R}^d$ is the vector of design variables, $\textbf{f}_{\text{i}}: \mathbb{R}^{\text{d}}\rightarrow\mathbb{R}^{m_i}$ is the multi-objective problem-specific function, $\textbf{g}_{\text{i}}: \mathbb{R}^{\text{d}}\rightarrow\mathbb{R}^{k_i}$ are the constraints to each sub-problem, $m_i$ and $k_i$ are the number of problem-specific objectives and constraints and $\text{i}\in\{\text{LO, TAS, MPS}\}$.

\subsection{Problem-specific assumptions}
\label{sub:3A_assumptions}
In order to reduce the computational complexity, we introduced some reasonable assumptions:

\begin{assumption}\label{assumption:static_planner}
    We considered the motion planner as \textit{static}: knowing the characteristics of the operations, the agents allocated and the resources, the trajectories are deterministically computed. Consequently, equation (\ref{equation:general_lower}) becomes $\textbf{x}_{\text{MP}}=\textbf{f}_{\text{MP}}(\textbf{x}_{\text{L}}, \textbf{x}_{\text{AS}})$, and this allowed us to remove equations (\ref{equation:general_lower}) and (\ref{equation:general_con3}) from the optimization loop.
\end{assumption}

\begin{assumption}\label{assumption:resources_availability}
    We introduced constraints and boundaries on LO in order to make all the resources always available to all the agents: with this assumption, the task allocation problem is no more a function of the layout.
\end{assumption}

\begin{assumption}\label{assumption:layout_and_scheduling_together}
    We separated the TAS problem into the sequential optimization of task allocation (TA) and task scheduling (TS): considering that the optimal TA is no more a function of LO, we decided to optimize them by defining the vector $\textbf{x} = [\textbf{x}_{\text{L}}^T, \textbf{x}_{\text{A}}^T]^T$, where $\textbf{x}_{\text{A}}$ contains the allocation variables, while the scheduling designs are stored inside a new vector called $\textbf{x}_{\text{S}}$.
\end{assumption}
Hence, the problem we aim at optimizing can be seen as the composition of two sub-problems, namely the simultaneous optimization of layout and task allocation (leader) and the optimization of the task scheduling (follower):
\begin{mini!}|s|[2]
    {\textbf{x}} 
    {\textbf{f}_{\text{LA}}(\textbf{x}, \textbf{x}_{\text{S}}) \label{equation:redefined_upper}} 
    {\label{equation:redefined_complete_equation}} 
    {} 
    \addConstraint{\textbf{g}_{\text{LA}}(\textbf{x}, \textbf{x}_{\text{S}}) \leq \textbf{0}}{\label{equation:redefined_con1}} 
    \addConstraint{\textbf{x}_{\text{S}} \in \min_{\textbf{x}_{\text{S}}} \textbf{f}_{\text{s}}(\textbf{x}, \textbf{x}_{\text{S}})}{\label{equation:redefined_lower}} 
    \addConstraint{\text{s.t.} \quad \textbf{g}_{\text{S}}(\textbf{x}, \textbf{x}_{\text{S}}) \leq \textbf{0}}{\label{equation:redefined_con2}} 
\end{mini!}

\subsection{Definition of agents, operations and tasks}
\label{sub:3B_operations_tasks}
The operations to be executed are contained in $\mathcal{O} = \{o_1, o_2, \cdots, o_t\}$ and they are defined by the precedence graph $\mathcal{G} = (\mathcal{O}, \mathcal{E})$. $\mathcal{A} = \{a_1, a_2, \cdots, a_n\}$ is the set of variables $a_i = i$, $\forall i\in\{1, 2, \cdots, n\}$ defining the actors involved in the process. We suppose to have full knowledge of the capabilities of the agents, encoded in a matrix $B_{n\times t}$, whose entries $b_{ij}\in\{\text{0,1}\}$ give information about the capability ($b_{ij} =$ 0) or non-capability ($b_{ij} =$ 1) of agent $a_i$ to perform operation $o_j$. For each operation $o_i$, the set $\mathcal{S}_i=\{s_1, s_2, \cdots,s_{d_i}\}=\{a_j\in\mathcal{A}\ |\ b_{j,i} \text{ = 0}\}, j=\{1,2,\cdots,n\}$ contains the indices of the $d_i$ agents capable of performing the $i$-th operation, $\mathcal{M} = \{m_1, m_2, \cdots, m_t\}$ is the set prescribing the exact number of agents required by each operation, while $\mathcal{I}\subseteq{\mathcal{O}}$ and  $\mathcal{C}\subseteq{\mathcal{O}}$ are the set storing the $v$ \textit{individual} operations and the $t-v$ collaborative ones. For us, each individual operation is associated to only one of the $l$ \textit{tasks} specified in $\mathcal{T}=\{t_1,\cdots,t_l\}$, whose definition depends on the HRC application under exam. For the operations in $\mathcal{C}$, we suppose to have full knowledge of the sequencing of the tasks composing them. We define two sets called $\mathcal{P}_r$ and $\mathcal{P}_h$, which contain respectively all the robotic \textit{elementary primitives} and the human \textit{atomic actions} needed by the static planner to create instances of the tasks in the simulation environment. Each task $t$, is characterized by a set $\mathcal{P}_{i,r}\subseteq\mathcal{P}_{r}$ and a set $\mathcal{P}_{i,h}\subseteq\mathcal{P}_{h}$ that contain the task-specific primitives/actions.
We suppose that we have only two types of agents: the first $\bar{h}$ elements of $\mathcal{A}$ and $\mathcal{R}$ (defined in Subsection \ref{sub:3C_resources}) are humans, while the subsequent $\bar{r}$ are robots. Consequently, we have that $n = \bar{h} + \bar{r}$. 

\subsection{Definition of the resources and the optimization variables}
\label{sub:3C_resources}

\begin{figure}
\vspace{0.2 cm}
\centering
\includegraphics[width=0.3\textwidth]{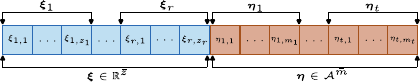}
\caption{Optimization vector for the \textit{leader} problem. \vspace{-0.5 cm}}
\label{fig:chromosome}
\end{figure}

We consider a set of resources, comprising all the agents and the pieces to be located in the environment, that we denote as $\mathcal{R} = \{r_1, r_2, \cdots, r_r\}$, while $\mathcal{Z}=\{z_1, z_2,\cdots,z_r\}$ contains the associated number of degrees of freedom (DOFs). Each resource is characterized by a feature vector $\boldsymbol{\Phi}_i = [\boldsymbol{\xi}_i, \boldsymbol{\gamma}_i]$, where $\boldsymbol{\xi}_i = [\xi_{i,1}, \xi_{i,2}, \cdots, \xi_{i,{z_i}}]$ represents the vector of coordinates to be optimized, while $\boldsymbol{\gamma}_i = [\textbf{v}_i^1, \textbf{v}_i^2, \cdots, \textbf{v}_i^{p_i}]^T$ contains the $p_i$ vertices, characterizing the parallelepiped with which we approximate each resource. Each couple of vertices $\textbf{v}_i^k$ and $\textbf{v}_i^{k+1},\ k=1,2,\cdots,p_i$, defines an edge $\textbf{e}_i^k = \textbf{v}_i^{k+1} - \textbf{v}_i^k = [\Delta{x_i^k},\ \Delta{y_i^k}]^T$ to which a normal vector $\hat{\textbf{n}}_i^k = [-\Delta{y_i^k}^T,\ \Delta{x_i^k}]^T$ is associated (see Figure \ref{fig:optimal_layout}b). We define with $\mathcal{N} = \{\hat{\textbf{n}}_{n + 1}^1,\cdots,\hat{\textbf{n}}_r^{p_r}\}$ the set containing all the $\sum_{i=n + 1}^r{p_i}$ normal vectors. We consider that the first two variables $\xi_{1,i}$ and $\xi_{2,1}$ of each vector $\boldsymbol{\xi}_i$ represent the coordinates$\{x_i,y_i\}$ of the centroid of the $i$-th resource, and we can define a vector $\textbf{p}_i=[\xi_{i,1}, \xi_{i,2}]^T$ that contains them. Consequently, regardless of the physical meaning of the optimization variables selected, we define $\boldsymbol{\xi} = [\boldsymbol{\xi}_1, \boldsymbol{\xi}_2, \cdots, \boldsymbol{\xi}_{r}] \in\mathbb{R}^{\bar{z}}$ , with $\bar{z} = \sum_{i=1}^r z_i$, as the overall design vector for the layout optimization part. Instead, we define the allocation vector as $\boldsymbol{\eta} = [\boldsymbol{\eta}_1, \boldsymbol{\eta}_2, \cdots, \boldsymbol{\eta}_t]\in\mathcal{A}^{\bar{m}}$, where $\bar{m}=\sum_{i=1}^t{m_i}$ and each $\boldsymbol{\eta}_i\in\mathcal{A}^{m_i}$ contains the indices (or a single index, in case $m_i$ = 1) of the agents that perform the $i$-th operation.

%% file: Chapters/4.Methodology.tex
\section{Methodology}
\label{sec:methodology}

\subsection{Leader problem: optimizing layout and task allocation}
\label{sub:4A_leader_redefinition}

The goal of the leader is to determine the set of variables to be used by the follower. In this work we introduce $m$ hyper-parameters contained in the vector $\textbf{w} = [w_1, w_2, \cdots, w_m]^T$ that allow to define a scalar objective function, calculated as the sum of the weighted objectives (KPIs). To cope with the different units of measure and order of magnitude of the $i$ KPIs, each of them is normalized with respect to $\bar{\mu}_i$ and $\bar{\sigma}_i$, obtained after creating the baseline, as explained in Section~\ref{sec:case_study}. Given the complexity of the problem we target, the relation between optimization variables and KPIs is unknown. We indicate as $\textbf{x} = [\boldsymbol{\xi}, \boldsymbol{\eta}]$ the overall optimization vector for the leader problem containing $\bar{d} = \bar{z} + \bar{m}$ heterogeneous variables (Figure \ref{fig:chromosome}), with $\boldsymbol{\sigma}(\textbf{x})$ the optimization vector for the scheduler (described in Subsection~\ref{sub:4B_follower_problem}) and we include all the $m$ normalized KPIs inside $\textbf{F}(\textbf{x},\boldsymbol{\sigma}(\textbf{x})) = [F_1, F_2, \cdots, F_m]^T$. The leader problem in equation~\ref{equation:redefined_complete_equation} becomes:
\begin{mini!}|s|[3]                   
    {\textbf{x}}                               
    {\textbf{f}(\textbf{x},\boldsymbol{\sigma}(\textbf{x})) = \textbf{w}^T\cdot\textbf{F}(\textbf{x},\boldsymbol{\sigma}(\textbf{x})) \label{eq:upper_level_minimization}}   
    {\label{equation:upper_level_optimization}}
    {}                               
    \addConstraint{\eta_{i,j} \in \mathcal{S}_i,\ \forall i\in\mathcal{I},\  j=1,\cdots,m_j}{\label{equation:upper_level_con1}}
    \addConstraint{\eta_{i,j} = s_j,\ \forall i\in\mathcal{C},\ j=1,\cdots,m_j}{\label{equation:upper_level_con2}}
    \addConstraint{\sum_{i=1}^v{\delta_{i,j}}\leq k_j,\ k_j\leq v,\ \forall i\in\mathcal{I},\ \forall j\in\mathcal{A}}{\label{equation:upper_level_con3}}
    \addConstraint{\bar{\xi}_{k,min} \leq \xi_k \leq \bar{\xi}_{k,max},\ k= 1,\cdots,\sum_{i=1}^r{z_i}}{\label{equation:upper_level_con4}}
    \addConstraint{
    \begin{aligned}
        d_{min,j}\leq \lVert \textbf{p}_i - \textbf{p}_j \rVert \leq d_{max,j},\ \forall j\in\mathcal{A},\\ 
        \forall i \in\{\bar{n}+1,\cdots,r\}
    \end{aligned}
}{\label{equation:upper_level_con5}}
\addConstraint{\bar{\mathcal{N}}\neq\emptyset,\ \forall s,q \in\{\bar{n}+1,\cdots,r\},\ s\neq q}{\label{equation:upper_level_con6}}
\end{mini!}

\begin{figure}
\vspace{0.2 cm}
\centering
\includegraphics[width=0.35\textwidth]{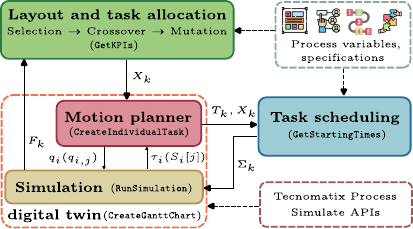}
\caption{Block diagram of the optimization framework: every block contains the function that is called in the recursive process, while the continuous arrows represent the data flow. \vspace{-0.8 cm}}
\label{fig:opt_framework}
\end{figure}

\noindent The constraint (\ref{equation:upper_level_con1}) imposes that the elements of the allocation vector $\boldsymbol{\eta}$ associated to an operation $\in\mathcal{I}$ are allocated to an agent that can perform the $i$-th operation. Constraint (\ref{equation:upper_level_con2}), on the contrary, is added to make sure that all the $m_j$ elements of the vector $\boldsymbol{\eta}_i$ are allocated to all the agents necessary for that $i$-th operation $\in\mathcal{C}$. Equation (\ref{equation:upper_level_con3}) is the last constraint on the allocation variables, and it allows to control the task distribution by limiting the number of tasks that can be allocated to an agent. More precisely, for all the $v$ \textit{individual} operations, each agent cannot be assigned to more than $k_j$ operations (the value of $\delta_{i,j}$ = 1 if operation $i$ is allocated to agent $j$, 0 otherwise). The three remaining constraints are for the layout. Equation (\ref{equation:upper_level_con4}) allows to limit the search range for each DoF (rectangles in Figure \ref{fig:optimal_layout}a). Instead, the constraint encoded in (\ref{equation:upper_level_con5}) imposes that the centroid of each resource is always placed inside the working areas of each agent, defined with maximum and minimum radii $d_{max,j}$ and $d_{min,j}$. The final constraint enforces the \textit{Separation Axis Theorem} (SAT)~\cite{huynh2009separating}, and it prevents the overlapping between the resources as shown in Figure \ref{fig:optimal_layout}b. For two generic resources $s$ and $q$, the requirement $\bar{\mathcal{N}}\neq\emptyset$ is verified only if at least one of the vectors $\hat{\textbf{n}}_{\alpha}^{\beta}$ contained in $\mathcal{N}$ is such that one between equation (\ref{equation:SAT_1}) and (\ref{equation:SAT_2}) holds, with $\alpha\in\{s,\ q\}$, $\beta\in\{1,2,\cdots,\text{max}(p_s,\ p_q)\}$, $j=1,2,\cdots,p_s$ and $k=1,2,\cdots,p_q$:
\begin{align}
\max_s \textbf{v}_s^j\cdot\frac{\hat{\textbf{n}}_{\alpha}^{\beta}}{\lVert \hat{\textbf{n}}_{\alpha}^{\beta} \rVert} < \min_q \textbf{v}_q^k\cdot\frac{\hat{\textbf{n}}_{\alpha}^{\beta}}{\lVert \hat{\textbf{n}}_{\alpha}^{\beta} \rVert} \label{equation:SAT_1}\\ 
\max_q \textbf{v}_q^k\cdot\frac{\hat{\textbf{n}}_{\alpha}^{\beta}}{\lVert \hat{\textbf{n}}_{\alpha}^{\beta} \rVert} < \min_s \textbf{v}_s^j\cdot\frac{\hat{\textbf{n}}_{\alpha}^{\beta}}{\lVert \hat{\textbf{n}}_{\alpha}^{\beta} \rVert} \label{equation:SAT_2}
\end{align}

\subsection{Follower problem: task scheduling}
\label{sub:4B_follower_problem}
The goal of the lower level optimizer is to find the vector $\boldsymbol{\sigma} = [\sigma_1, \sigma_2, \cdots, \sigma_t]^T$, containing the optimized starting times for each operation. To do this, we aim at minimizing the maximum completion time $c_i$ of the last scheduled task. However, it must be noted that the input to the scheduler is not directly the vector $\textbf{x}$, but the vector $\boldsymbol{\tau}(\textbf{x})=[\tau_1, \tau_2, \cdots, \tau_t]^T$, which contains the cycle time taken by the agents, allocated according to $\boldsymbol{\eta}$, to perform the $t$ operations. In the end, it is straightforward to say that $c_i=\sigma_i+\tau_i(\textbf{x})$, since the cycle times contained in $\boldsymbol{\tau}$ have already been evaluated for the overall operation.\\
The collaborative process is described by the precedence graph $\mathcal{G}$ defined in Subsection~\ref{sub:3B_operations_tasks}. Considering $\mathcal{O}$ as the set of nodes (operations) and $\mathcal{E}=\mathcal{O}\times\mathcal{O}$ as the set of directed edges, the precedence information contained in $\mathcal{G}$ is encoded into matrix $P_{t\times t}$, characterized by entries $P_{ij}\in\{\text{0,1}\}$, $\forall i,j\in \{1, 2. \cdots, t\}$ specifying if an operation $o_i$ precedes an operation $o_j$ ($P_{ij}$ = 1) or if it does not ($P_{ij}$ = 0). Moreover, to keep track of the agents availability, the measure $R_k$ is introduced, and it is constantly updated with the time at which each agent $k$ becomes available.
The lower-level optimization problem becomes:
\begin{mini!}|s|[3]                   
    {\boldsymbol{\sigma}}                               
    {\max_{i\in\{1,2,\cdots,t\}} c_i \label{eq:lower_level_minimization}}   
    {\label{equation:lower_level_optimization}}
    {}                               
    \addConstraint{\sigma_j\ \geq\ 0,\ \forall{j\ \in\ \{1,2,\cdots,t\}}}{\label{equation:lower_level_con1}}
    \addConstraint{
    \begin{aligned}
        \sigma_j\ \geq\ \max ({\max_{i\in\{i\ |\ P_{ij} = 1\}} ({\sigma_i + \tau_i})}, \\
        \max_{k\in\{\eta_{j,q}\ |\ q=1,\cdots,m_j\}}({R_k}))
    \end{aligned}
    }{\label{equation:lower_level_con2}}
    \addConstraint{
    \begin{aligned}
        R_k=\sigma_i+\tau_i,\ \forall{k}\in\{\eta_{i,j}\ |\ j=1,\cdots,m_i\}, \\
        \forall{i}\in\{1,2,\cdots,t\} 
    \end{aligned}
    }{\label{equation:lower_level_con3}}
\end{mini!}

The constraint (\ref{equation:lower_level_con1}) states that all starting times must be positive for each considered operation. In equation (\ref{equation:lower_level_con2}), the generic $j$-th operation is imposed to start after all its predecessors are completed (all the $i$ tasks such that $P_{ij}$ = 1) and also after that all the $k$ agents required are available (these $m_j$ agents are represented by the $q$ elements of vectors $\boldsymbol{\eta}_j$). Finally, constraint (\ref{equation:lower_level_con3}) represents the update rule for the availability metric $R_k$.

\subsection{Description of the genetic algorithm}
\label{sub:4C_genetic_algorithm}
To solve equation (\ref{equation:upper_level_optimization}) we devised a genetic algorithm, whose stages are sequenced according to the mainstream framework presented in~\cite{lambora2019genetic}. The whole procedure requires to generate $N_p$ random parent chromosomes by enforcing the constraints in equation (\ref{equation:upper_level_optimization}), in order to fill the $(N_p \times \bar{d})$ matrix $X_p$. The vector $\textbf{f}_k$ contains the evaluation of the fitness function for the $N_k$ \textit{individuals}, that can be both parents ($k=p$) or children ($k=c$) as a function of the specific step. After the the parents generation, the recursive procedure is started: at the first iteration, the algorithm \texttt{GetKPIs} (see Algorithm \ref{alg:get_KPIs}) allows to retrieve the $m$ KPIs for each of the $N_p$ parents and store them in a $(m \times N_p)$ matrix $F_p$. Based on $F_p$ it is possible to evaluate the fitness of each parent chromosome and obtain the $N_p$-dimensional vector $\textbf{f}_p$, according to equation (\ref{equation:upper_level_optimization}). In order to generate the $N_c$ children to be tested, \textit{Selection}, \textit{Crossover} and \textit{Mutation} are executed recursively, to make sure the constraints are respected. The parents selection is based on the \textit{Roulette wheel} approach~\cite{anand2015novel}: two numbers between 0 and 1 are randomly generated, and the two parents whose costs are closer to them are added to the ($N_c \times \bar{d}$) matrix $X_p^*$. This process is repeated $N_c/2$ times, and the cost $\pi_i$ for each individual (such that $\sum_{i=1}^{N_p}{\pi_i} = 1$) is calculated by leveraging the \textit{Boltzmann} distribution, after selecting an hyper-parameter $\beta$:
\begin{equation}
\pi_i^*=exp({-\beta \frac{\textbf{f}_p[i]}{\frac{1}{N_p}\sum_{j=1}^{N_p}{\textbf{f}_p[i]}}}),\ \pi_i = \frac{\pi_i^*}{\sum_{j=1}^{N_p}{\pi_i^*}}
\label{equation:selection_Boltzmann}
\end{equation}
The crossover phase varies as a function of the nature of the variables in the chromosomes. For real variables, a vector $\textbf{v}_j=[v_{1,j},\cdots,v_{\bar{z},j}]$ composed of 0s and 1s is created, and the $N_c/2$ couples of chromosomes required can be calculated as the element-wise product between the $j$ rows of $X_p^*$ (up to the $\bar{z}$-th element) and $\textbf{v}_j$ summed to the product between the same set of elements and the reciprocal (1 - $\textbf{v}_j$). 
\noindent For the discrete designs, each $k$-th vector $\boldsymbol{\eta}_k$ of the rows $j$ and $j+1$ ($k=\{1,\cdots,t\}$) is randomly selected according to equation (\ref{equation:upper_level_con1}), $\forall{k}\in\mathcal{I}$, or it is forced to satisfy (\ref{equation:upper_level_con2}), $\forall{k}\in\mathcal{C}$.\\
For each individual in $X_c$, a vector with the same length as each chromosome is generated and filled with random numbers between 0 and 1: $\forall{i}=\{1,\cdots,N_c\}$, the indices of the elements of each $i$-th chromosome having a value smaller than the mutation rate $\mu_0$ are saved in the set $\mathcal{K}_j$. For discrete variables, each element whose index is saved in $\mathcal{K}_j$ and is bigger than $\bar{z}$ is switched with another suitable value, enforcing equations (\ref{equation:upper_level_con1}) and (\ref{equation:upper_level_con2}). For real variables, each $k$-th element contained in $\mathcal{K}_j$ (and smaller than $\bar{z}$) becomes $X_c[i,k]=X_c[i,k] + {\sigma_0} \cdot Z_k$, with $Z_k \sim \mathcal{N}(0,1)$. At this point, the constraints are checked and, if some of them turn out to be \textit{non-valid}, the sequence starting from the \textit{Selection} phase described above is triggered again. This process continues until $N_c$ \textit{valid} children are contained in $X_c$.
\noindent Once the $N_c$ individuals are set, the $m$ KPIs for the children chromosomes are retrieved and the matrix $F_c$ is updated. At this point, the vector of fitness functions for the children $\textbf{f}_c$ is calculated and, after sorting the chromosomes in ascending order of fitness, the population of parents $X_p$ and corresponding fitness vector $\textbf{f}_p$ are respectively updated with the first $N_p$ individuals of $X_c$ and the same number of elements from $\textbf{f}_c$. 
Before starting a new iteration, a check on the best solution is performed: if any $k$-th element in $\textbf{f}_p$ is smaller than the current best fitness $f^*$, then $f^*$ is overwritten and also $\textbf{x}^*$ is changed with the corresponding $X_p[k,:]$. Finally, to avoid the \textit{stagnation} of $f^*$, the process of $\textit{Adaptive mutation}$ takes place: in case that $f^*$ has not been updated for more than $\bar{N}_{it}$=2 consecutive iterations, $\mu_0$ becomes $1.05 \cdot \mu_0$, while $\sigma_0$ is reduced to $0.95 \cdot \sigma_0$  (more elements of the chromosomes mutate, but the change is less evident). As soon as $f^*$ is updated, $\mu_0$ and $\sigma_0$ are changed to the original values.

\subsection{Digital model and task scheduler}
\label{sub:4D_motion_planner}
As explained in assumption~\ref{assumption:static_planner}, we consider the motion planner to be \textit{static}. From the leader, a matrix of individuals $X_k$ (with $k=\{p,c\}$) is passed to the function \texttt{GetKPIs}, that exploits the APIs of the simulator Tecnomatix Process Simulate (TPS) by Siemens to run the simulations and return the matrix of KPIs $F_k$. First of all, $X_k$ is processed by the \textit{state machine} that represents our motion planner. By calling \texttt{CreateIndividualTask}, the algorithm gets access to the task $t_i$ associated to the $i$-th operation under exam and to the set $\mathcal{P}_{i,z}$, with $z\in\{h,r\}$ as a function of the information stored in $X_k$, and creates the instance of the task, called $q_i$, in the simulation environment. In case $i\in{\mathcal{I}}$, the algorithm checks if the variable $\eta_i$ takes values between 1 and $\bar{h}$: in case it does, the operation to be created is for the human agent whose index is stored in $\eta_i$. Otherwise, the $i$-th operation will be for a robot. Instead, if $i\in\mathcal{C}$, more tasks $q_{i,j}$ are needed: the planner scans the vector $\boldsymbol{\eta}_i$, containing $m_i$ elements, and repeats the procedure described above. For operations in $\mathcal{C}$, the function \texttt{CreateGanttChart} allows to take all the $m_i$ instances of the tasks $q_{i,j}$ created (saved in a vector $Q_i$) and join them in the prescribed sequence (we suppose to know the sequencing, as explained in~\ref{sub:3B_operations_tasks}: based on the times $S_i$ required by the tasks, the starting times for the $j$ tasks are known). When an instance $q_i$ is created, the planner updates the vector $\boldsymbol{\tau}$ storing the cycle times. Since the genetic algorithm produces $N_k$ individuals to be tested, the planner repeats the evaluation of the cycle times for $N_k$ times, and then it returns the matrix $T_k$.\\
The function \texttt{GetStartingTimes} triggers the task scheduler, external to the simulator. Based on the cycle times contained in $T_k$ for the $N_k$ individuals represented by $X_k$, This method allows to obtain the ($N_k \times t$) matrix $\Sigma_k$ by solving the optimization problem (\ref{equation:lower_level_optimization}). $\Sigma_k$ contains the starting times (denoted with $\boldsymbol{\sigma}$ in equation (\ref{equation:lower_level_optimization})) for the $N_k$ individuals. In order to create and simulate the complete operation, \texttt{CreateGanttChart} schedules the previously created $Q_k$ instances of the operations and runs the simulations to obtain the KPIs. The procedure described in this Subsection and in~\ref{sub:4C_genetic_algorithm}, schematized in Figure \ref{fig:opt_framework}, is repeated for $N_{it} + 1$ iterations, that correspond to $N_c \cdot N_{it} + N_p$ simulations ($N_s$ in Table \ref{table:param}) in TPS. Algorithm~\ref{alg:get_KPIs} is the C\# \textit{pseudo code} that exploits the APIs to create, at each recursion, virtual objects needed in the specific simulation, with the aim of evaluating KPIs that are representative of the overall HRC process.

\begin{algorithm}
\scriptsize
\caption{\texttt{GetKPIs}}
\label{alg:get_KPIs}
\begin{algorithmic}[1]

    \STATE \textbf{Given}: $X_k$
    \STATE $N_k \gets$ number of rows of $X_k$, $Q_k \gets$ empty vector of instances
    \FOR{$a \gets 1,\dots, N_k$}
        \FOR {$i \gets 1,\dots, t$}
            \IF{$i\in{\mathcal{I}}$}
                \STATE $q_i \gets$ \texttt{CreateIndividualTask}($t_i$, $\mathcal{P}_{i,z}$)
                \STATE $\boldsymbol{\tau}[i] \gets$ \texttt{RunSimulation}, $Q_k[i] \gets q_i$
            \ELSIF{$i\in{\mathcal{C}}$}
            \STATE $Q_i \gets$ \textbf{0}
                \FOR{$j \gets 1,\dots, m_i$}
                    \STATE $q_{i,j} \gets$ \texttt{CreateIndividualTask}($t_i$, $\mathcal{P}_{i,z}$)
                    \STATE $S_i[j] \gets$ \texttt{RunSimulation}, $Q_i[j] \gets q_{i,j}$
                \ENDFOR
                \STATE $q_i \gets$ \texttt{CreateGanttChart}($Q_i$, $S_i$), $Q_k[i] \gets q_i$
                \STATE $\boldsymbol{\tau}[i] \gets$ \texttt{RunSimulation}
            \ENDIF
        \ENDFOR
        \STATE $T_k[a, :] \gets \boldsymbol{\tau}$
    \ENDFOR
    \STATE return $T_k$, $Q_k$
    \STATE $\Sigma_k \gets$ \texttt{GetStartingTimes}($T_k$, $X_k$)
    \STATE \text{void}$\gets$\texttt{CreateGanttChart}($Q_k$, $\Sigma_k$), $F_k \gets$\texttt{RunSimulation}
    \STATE return $F_k$

\end{algorithmic}
\end{algorithm}
\vspace{-0.6cm}

%% file: Chapters/5.Experiment.tex
\section{Case study}
\label{sec:case_study}

\begin{figure}
\vspace{0.1 cm}
\centering
\vspace*{+0.1 cm}
\includegraphics[width=0.4\textwidth]{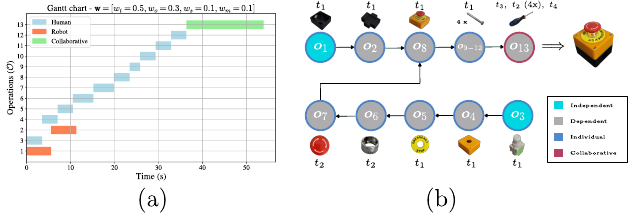}
\caption{(a) Optimal scheduling corresponding to the best solution in Figure~\ref{fig:optimal_layout}a. (b) Precedence graph of the emergency stop button: we assume that $o_{9-12}$ can only be executed by the human, while $o_{13}$ is collaborative. 
\vspace{-0.4 cm}
}
\label{fig:Gantt_precedence}
\end{figure}

\begin{figure}
\centering
\vspace*{-0.2cm}
\includegraphics[width=0.44\textwidth]{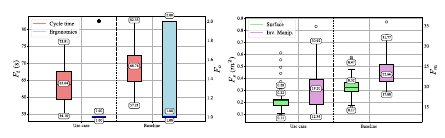}
\caption{Data distributions for the metrics considered in the test, compared to the baseline. The labels represent the minimum, maximum and median values. 
\vspace{-0.5 cm}
}
\label{fig:box_plots}
\end{figure}

The proposed framework is validated in simulation, since it was engineered to be applied in the pre-deployment phase. A Universal Robots UR5e cobot, equipped with an electro-mechanical Robotiq Hand-E gripper was considered. We tested our framework in the assembly of the emergency stop button shown in Figure~\ref{fig:Gantt_precedence}b, and the main parameters we selected are reported in Table~\ref{table:param}. The planner is able to create the instances of each $i$-th task $t_i$ by using the task-specific robotic \textit{primitives} ($\mathcal{P}_{i,r}$) and human \textit{actions} ($\mathcal{P}_{i,h}$) contained in $\mathcal{P}_r=\{\textit{MoveTo},\ \textit{Overfly},\ \textit{Screwing},\ \textit{Open},\ \textit{Close}\}$ and $\mathcal{P}_h=\{\textit{Get},\ \textit{Put},\ \textit{Pose},\ \textit{Wait}\}$, as shown in the video.\\
With our methodology, we expected to obtain a solution that attains better performances than an initial random one. For this reason, we created a baseline for the comparisons by forcing the fitness values contained in $\textbf{f}_p$ to be equal to 0: with this expedient, the $N_p$ parents selected before starting a new iteration are chosen in a random way and also each parent has the same probability to be selected at the following iteration (equation (\ref{equation:selection_Boltzmann})). We considered this process to be representative of the choices of 124 people selecting how to execute the assembly operation, respecting boundaries and constraints. The resulting KPIs allowed us to evaluate the values of $\bar{\mu}_i$ and $\bar{\sigma}_i$ mentioned in Subsection~\ref{sub:4A_leader_redefinition}.\\
We selected four normalized objectives: cycle time ($F_t$), ergonomics ($F_o$), inverse manipulability ($F_m$) and minimum surface ($F_s$). Moreover, we added a flag to the summation: to account for safety: the flag $t_s$ is set to $\infty$ if a collision happens, otherwise its value is 0. To evaluate $F_o$, the OWAS score was considered~\cite{karhu1981observing}: at each discretization step $\Delta_t$ of the simulation the four needed indices are evaluated and we considered their average values to obtain the final score from the table. For the manipulability measure, we selected $F_m = 1/{\det(J)}_{mean}$, where ${\det(J)}_{mean}$ is the mean among the registered values for each simulation.\\ 
The dispersion of the data characterizing our test is reported in Figure~\ref{fig:box_plots}, where the KPIs are shown with their correct unit of measure. We selected $\textbf{w}=[0.5, 0.3, 0.1, 0.1]$ since we wanted to prioritize the reduction of the cycle time, in order to maximize the throughput of the cell. The second KPI for importance is the ergonomics.
The median of the cycle time (63,64) is smaller compared to the baseline (68,76) and the cycle time characterizing the highest value of $f^*$ (57.58) is just slightly higher than the minimum cycle time of the baseline (57.21): the weight $w_t$=0.5 guarantees to obtain a significant cycle time reduction. The ergonomics metric presents the same median as the randomic generation. From the baseline, we see that the median is equal to 1, while the maximum is 2 (and there are no outliers); therefore, it's possible to say that all the simulations already resulted in an ergonomic score that was either 1 or 2, which is acceptable when dealing with the OWAS. In the test, the weight $w_0$=0.3 is enough to minimize $F_o$ for the majority of the simulations (the interquartile range is null), since the values equal to 2 are outliers and not so representative of the real distribution of the data. For $F_s$ and $F_m$, despite they were given the smallest weights ($w_s=w_m$=\ 0.1), the algorithm allowed to decrease both the indicators. As it can be seen from the higher number of outliers though, these weights are usually not enough to obtain a steady decrease in the values of these KPIs: especially for $F_m$, this can be explained by the presence of singularities inside the working volume, whose occurrence is difficult to predict. It must be noted that the data presented in Figure~\ref{fig:box_plots} represent the 124 simulations; however, as explained in Subsection~\ref{sub:4C_genetic_algorithm}, the genetic algorithm arrives to the best solution by comparing each child with the current best. From the Gantt chart (Figure \ref{fig:Gantt_precedence}a), it is possible to see that the algorithm is suggesting the expected optimal allocation $\boldsymbol{\eta}$. The suggestion is to parallelize $o_1\rightarrow o_2$ and $o_3\rightarrow o_4$, while the remaining operations (excluding $o_{9-12}$ and $o_{13}$ that can only be executed in a specific way in our case study) are allocated to the human, who is always faster than the robot. The final configuration can be seen in Figure~\ref{fig:optimal_layout}a. The positions of $r_3$ and $r_4$, where the robot picks the pieces during $o_1$ and $o_2$, do not influence the overall cycle time: as expected, since $w_s<w_t$, the final result does not present the smallest surface, and this is visible by the locations of $r_3$ and $r_4$ that are quite far from the robot. The robot is put as close as possible to $r_{10}$: as it can be seen from the accompanying video, this allows to put the pieces in $r_9$ with ease, not reducing $F_m$ too much. Also, the fixtures $r_{5-6-7-8}$ are correctly placed as close as possible to the human.

\begin{table}
\vspace{0.4 cm}
    
    \centering 
    \tiny
    \begin{tabular}{|c|c|c|c|c|c|c|c|c|c|c|}
    \hline
    \rowcolor[HTML]{EFEFEF}  
     $t$ & $n$ & $z_i$ & $m$ & $\mu_0$ & $\sigma_0$ & $\Delta_t$ & $N_p$ & $N_c$ & $N_{it}$ & $N_s$ \\
    \hline
     13  & 2  & 2  & 4  & 0.25  & 100  & 0.01  & 4 & 6   & 20  & 124 \\
    \hline 
    \end{tabular}
    \caption{User-defined values. }
    \label{table:param}
\end{table}

%% file: Chapters/6.Conclusion.tex
\section{Conclusions and future works}
\label{sec:conclusions}
This paper proposes a novel approach for the optimization of HRC cells by utilizing a simulator during the pre-deployment phase. The proposed black-box recursive method can accomodate for production and human-centric KPIs to evaluate the layout, task allocation and scheduling, accounting for a binary logic for safety. The case study shows the potential of the method in the deployment of HRC cells. Future research will focus on introducing a dynamic planner in the optimization framework and evolving the pipeline to multi-objective settings, in order to account for an increased amount of variables and to have a more robust baseline. 